\documentclass[english]{llncs}
\usepackage{lmodern}
\usepackage[T1]{fontenc}
\usepackage[latin9]{inputenc}
\usepackage{float}
\usepackage{units}
\usepackage{stackrel}
\usepackage{graphicx}

\makeatletter

\providecommand{\tabularnewline}{\\}

\usepackage{graphicx}
\usepackage[font={small}]{caption}

\@ifundefined{showcaptionsetup}{}{%
 \PassOptionsToPackage{caption=false}{subfig}}
\usepackage{subfig}
\makeatother

\usepackage{babel}
\begin{document}

\title{Demographic-Guided Attention in Recurrent Neural Networks for Modeling
Neuropathophysiological Heterogeneity}

\author{Nicha C. Dvornek\textsuperscript{1,2}, 
Xiaoxiao Li\textsuperscript{2}, 
Juntang Zhuang\textsuperscript{2}, 
Pamela Ventola\textsuperscript{3}, 
and James S. Duncan\textsuperscript{1,2,4,5}} 

\institute{\textsuperscript{1}Radiology \& Biomedical Imaging,
Yale School of Medicine, New Haven, CT, USA\\
\textsuperscript{2}Biomedical Engineering, Yale University, New Haven, CT, USA\\
\textsuperscript{3}Child Study Center,
Yale School of Medicine, New Haven, CT, USA\\
\textsuperscript{4}Electrical Engineering, Yale University, New Haven, CT, USA\\
\textsuperscript{5}Statistics
and Data Science, Yale University, New Haven, CT, USA}

\maketitle
\begin{abstract}
Heterogeneous presentation of a neurological disorder suggests potential
differences in the underlying pathophysiological changes that occur
in the brain. We propose to model heterogeneous patterns of functional
network differences using a demographic-guided attention (DGA) mechanism
for recurrent neural network models for prediction from functional
magnetic resonance imaging (fMRI) time-series data. The context computed
from the DGA head is used to help focus on the appropriate functional
networks based on individual demographic information. We demonstrate
improved classification on 3 subsets of the ABIDE I dataset used in
published studies that have previously produced state-of-the-art results,
evaluating performance under a leave-one-site-out cross-validation
framework for better generalizeability to new data. Finally, we provide
examples of interpreting functional network differences based on individual
demographic variables.
\end{abstract}

\section{Introduction}

Functional magnetic resonance imaging (fMRI) has begun to play a large
role in characterizing the neuropathophysiology of psychiatric disorders.
One example is in the characterization of autism spectrum disorder
(ASD), a neurodevelopmental disorder that affects communication and
behavior. ASD is extremely heterogeneous, presenting with a wide range
of symptoms and severity of impairments. Early fMRI studies investigated
small datasets with imposed homogeneity, e.g., restricting to one
gender, age group, or level of functioning. However, this resulted
in smaller datasets, largely irreproducible results and lack of generalization
to new datasets. More recently, the popular large public Autism Brain
Imaging Data Exchange (ABIDE) I resting-state fMRI dataset \cite{DiMartino2014}
has undergone extensive analysis, including the application of machine
learning to classify ASD and healthy controls (HC) for the purpose
of discovering neuroimaging biomarkers of ASD. However, even with
the large amount of neuroimaging data, achieving high classification
performance has been a challenge, likely due in part to both the heterogeneity
of the sample populations of each imaging site and the heterogeneity
of the underlying neurological mechanisms of the disorder itself.
Evidence for these potential reasons includes the much poorer performance
of leave-one-site-out cross-validation (LOSO CV) compared to intrasite
k-fold cross-validation \cite{Abraham2017,Heinsfeld2018}. 

One approach to mitigating the heterogeneity is to incorporate demographic
information into the classification problem. Here, we refer to demographic
variables as non-imaging, scalar variables that are often measured
and easy to obtain, such as gender, age, or IQ. Demographic information
can be incorporated in different ways depending on the classification
model. For example, the demographic information can be fused at different
layers in a standard feedforward neural network \cite{Dvornek2018,ngiam2011}
or used as targets for prediction \cite{Dvornek2018}. Furthermore,
demographic information can be combined in model specific ways, e.g.,
to define the edges in graph-based models \cite{Parisot2017} or to
set the initial state of recurrent neural network models \cite{Karpathy2016,Dvornek2018a}.
However, none of these approaches aim to modulate the underlying neurological
differences that may be describing the heterogeneity in ASD.

To model disorder heterogeneity in terms of changes in the underlying
functional processing, we propose a demographic-guided attention module
to enhance a recurrent neural network model for processing fMRI time-series
data. While the attention scores are computed across time, we can
interpret the resulting context as guiding attention to different
functional networks. In addition to using the demographic information
to help identify which functional networks to attend to in classifying
ASD or HC, we propose a novel loss for computing more diverse queries
for each attention head to better model the sample heterogeneity.
We compare our proposed methods to other ways of handling demographic
data on 3 subsets of the ABIDE dataset, matched to previous studies
that have previously demonstrated state-of-the-art results from the
fMRI data alone. We achieve some of the highest accuracy of ABIDE
classification under LOSO CV. Finally, we give examples of functional
networks that may undergo diverse processing in ASD based on individual
demographic factors.

\section{Methods}

We build on recent models for predicting from fMRI time-series data
that use recurrent neural networks with long short-term memory (LSTM).
To model the heterogeneity of ASD, we apply a generalized attention
mechanism that is guided by individual demographic characteristics.
The context learned from the attention mechanism is then used to bias
the LSTM outputs, allowing the model to focus on different functional
networks based on individual non-imaging characteristics (Fig. 1). 

\begin{figure}

\centering{}\includegraphics[bb=0bp 125bp 700bp 505bp,clip,width=0.75\textwidth]{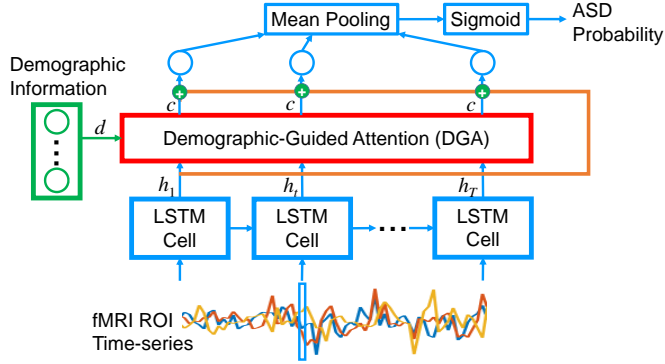}\vspace{-5bp}
\protect\caption{Demographic-guided attention network for classification of ASD/HC
from fMRI. }
\end{figure}

\subsection{Network Architecture}

\subsubsection*{Baseline LSTM for fMRI Time-Series}

The baseline LSTM network to predict from fMRI time-series was first
proposed by Dvornek et al. \cite{Dvornek2017}. The fMRI time-series
with length $T$ from regions of interest (ROIs) in a predefined brain
parcellation is first input to the LSTM layer. Then the output of
the LSTM cell at each timepoint $h_{t}\in R^{n}$ is input to a fully
connected (FC) layer with weights shared across time. The outputs
of the FC layer are averaged across time and input to a sigmoid activation
function to produce the probability of ASD label.

\subsubsection*{Demographic-Guided Attention}

We propose to incorporate functional network differences resulting
from disease heterogeneity through a generalized attention mechanism
\cite{Vaswani2017}. The attention mechanism can be described as a
function mapping a query and key-value pair to some output, often
referred to as the context or a head. In our work, the query is defined
by the demographic information, and the key and value are defined
by the outputs of the LSTM layer $h_{t}$. Applying the scaled dot
product attention \cite{Vaswani2017}, the context vector $c$ is
computed by
\begin{equation}
c=att\left(d,\left\{ h_{t}\right\} \right)=\stackrel[t=1]{T}{\sum}softmax\left[\nicefrac{\left(W_{q}d\right)^{T}\left(W_{k}h_{t}\right)}{\sqrt{m}}\right]W_{v}h_{t},
\end{equation}
where $d\in R^{l}$ is the vector of demographic information; $softmax\left[a_{t}\right]=\nicefrac{\exp\left(a_{t}\right)}{\stackrel[j=1]{T}{\sum}\exp\left(a_{j}\right)}$
with $a_{t}=\nicefrac{\left(W_{q}d\right)^{T}\left(W_{k}h_{t}\right)}{\sqrt{m}}$;
and $W_{q}\in R^{m\times l}$, $W_{k}\in R^{m\times n}$, and $W_{v}\in R^{m\times n}$
are weight matrices that operate on $d$ or $h_{t}$ to define the
query, key, and value vectors, respectively. In this work, we set
$m=n$.

\subsubsection*{Residual Connection for Modeling Heterogeneity}

In standard attention approaches, the context vector is concatenated
with the LSTM output \cite{Bahdanau2015} or the context vectors alone
\cite{Vaswani2017} are used as input to the following layers. Here,
we propose to use the context to bias the output of the LSTM layer,
changing the focus on LSTM nodes that should be emphasized based on
the demographic information. We do this by simply adding a residual
connection betwen the output of each attention head $k$ and the output
of the LSTM layer, $c_{k}+h_{t}$ (Fig. 1, orange path). The summed
outputs are then processed in a similar way as the baseline LSTM model,
using a FC layer with shared weights and averaging the FC outputs
over time. If multiple attention heads are used, then each head is
separately processed with a different FC layer, and the maximum score
across the heads is passed to the sigmoid layer to represent probability
of ASD. The rationale for keeping the maximum score is that only one
mode of functional network patterns may be indicative of ASD.

\subsection{Query Diversity Loss \label{sub:FC}}

A single attention module allows for attending to different LSTM nodes
based on the demographic information. However, this assumes then that
two individuals with the same demographic profile must share the same
underlying neuropathology. To allow for even greater diversity in
modeling disease heterogeneity, we can include more attention heads
that will learn different contexts. To encourage the different attention
heads to capture different underlying neuropathological modes, we
propose the following query diversity loss (QDL):

\begin{equation}
L_{QD}=\stackrel[i=1]{N}{\sum}\stackrel[j=1]{K-1}{\sum}\stackrel[k=j+1]{K}{\sum}\left|\frac{q_{ij}^{T}q_{ik}}{\left\Vert q_{ij}\right\Vert \left\Vert q_{ik}\right\Vert }\right|
\end{equation}
where $N$ is the number of subjects and $q_{ij}=W_{q_{j}}d_{i}$
is the $n$-dimensional query vector for attention head $j$. QDL
computes the cosine proximity for all query vectors $q_{ij}$ for
subject $i$. Minimizing QDL thus encourages projection of the demographic
information into orthogonal subspaces, which capture complementary
information, before comparing to the keys to compute the attention
scores.

The total loss $L$ is then 
\begin{equation}
L=L_{C}+\lambda L_{QD},\label{eq:loss}
\end{equation}
 where $L_{C}$ is the classification loss (e.g., binary cross-entropy)
and $\lambda$ is a hyperparameter controlling the contribution of
QDL.

\subsection{Interpretation of Attention as Neuropathological Heterogeneity}

We first interpret each node of the LSTM as modeling a functional
network. While different attribution methods can be applied, we follow
Dvornek et al. and assign ROIs to a network if the LSTM weights for
the ROI inputs have large magnitude (> 3 standard deviations above
 mean weight magnitude) \cite{Dvornek2017}.

The proposed model uses the context computed by the demographic-guided
attention module as a bias for the LSTM outputs. Since each node of
the LSTM is interpreted as processing the signal corresponding to
some functional network, we interpret the demographic information
as providing context for deciding which functional networks should
be given more attention in performing ASD classification, i.e., we
measure the demographic-guided attention to a functional network $f$
as $c\left(f\right)$. We then assess the coupling between a functional
network and a demographic variable by computing the correlation between
the demographic variable $d\left(i\right)$ and the context $c\left(f\right)$
for functional network $f$ across subjects. Different patterns of
attention for a functional network in different attention heads allows
for modeling greater neuropathological heterogeneity.

\section{Experiments}

\subsection{Data}

We use resting-state fMRI data from the multisite ABIDE I dataset
\cite{DiMartino2014} which was released by the Preprocessed Connectomes
Project \cite{Craddock2013}. To demonstrate robustness of our approach
and directly compare with results from the literature, we analyzed
the same subsets of data under the same preprocessing conditions as
in 3 prior studies: Dataset 1 (DS1) from \cite{Dvornek2017}, with
$N=1100$ subjects, preprocessed using the Connectome Computation
System pipeline, band-pass filtering and no global signal regression,
and parcellated with the CC200 atlas; Dataset 2 (DS2) from \cite{Heinsfeld2018},
with $N=1035$ subjects, preprocessed using the Configurable Pipeline
for the Analysis of Connectomes, band-pass filtering and global signal
regression, and parcellated with the CC200 atlas; and Dataset 3 (DS3)
from \cite{Abraham2017}, with $N=870$ subjects, preprocessed using
the same pipeline as in \cite{Heinsfeld2018} but parcellated with
the HO atlas. 

The time-series for each ROI of each subject was standardized by subtracting
the mean and dividng by the standard deviation and resampled to 2s
intervals between time points to harmonize the sampling across acquisition
sites. We augmented the dataset by a factor of 10 during training
by extracting 10 randomly cropped windows with length $T=90$ timepoints
from each subject during each epoch. At test time, every possible
window of 90 timepoints is extracted from the time-series data for
each subject and input to the trained network. The predicted probability
of ASD for a given subject was then computed as the proportion of
windowed samples classified as ASD. 

Demographic information included gender, age, handedness, full IQ,
verbal IQ, performance IQ, and eye status during scanning. Missing
IQ data were imputed based on other available IQ scores for the subject,
where we approximated full IQ as the average of verbal IQ and performance
IQ, and subjects with no available IQ scores were assigned scores
of 100, which is the mean population IQ. Each demographic variable
was standardized to lie in the range of {[}-1,1{]}.

\subsection{Experimental Methods}

Models for classification of ASD vs. HC were trained for each subset
of the ABIDE dataset. We compared and implemented the following models
which have the same underlying LSTM baseline architecture and incorporate
demographic information: the proposed demographic-guided attention
network (DGA); the DGA network without the residual connection, i.e.
using the computed context alone (DGA-C); the baseline LSTM network
combined with separately processed demographic information through
late fusion as proposed in \cite{Dvornek2018} (DFuse); the baseline
LSTM network with the hidden state and cell state of the LSTM initialized
based on the demographic information as proposed in \cite{Dvornek2018a}
(DInit). Models were implemented in Keras, with 32 nodes for the LSTM.
For regularziation, models were trained using a dropout layer before
each fully connected layer (with 0.5 probability of node dropout).
Optimization was performed using the Adam optimizer, with binary cross-entropy
loss or with QDL as in Eq. \ref{eq:loss} for DGA2, a batch size of
32, and early stopping based on validation loss and a patience of
5 epochs. DGA-based models were tested with 1 (DGA1) or 2 (DGA2) attention
heads and QDL with $\lambda=0.5$ (DGA2-QDL). In addition, we compared
the original study for each dataset that used only imaging information.

To assess the implemented models, we used LOSO CV, repeating the CV
5 times and averaging the performance measures for each site across
CV runs both with and without weighting by the number of subjects
per test site. We chose the LOSO framework to better estimate the
model generalizeability compared to the commonly employed stratified
k-fold cross-validation, which gives overoptimistic results. We measured
classification performance by computing the accuracy (ACC), true positive
rate (TPR), true negative rate (TNR), and area under the receiver
operating characteristic curve (AUC). We tested for differences against
the baseline LSTM model by comparing the performance for the same
left-out sites using two-tailed paired t-tests with a significance
level of 0.05.

We also evaluated functional networks that were attended to based
on individual demographic factors by applying the Neurosynth decoder
\cite{Yarkoni2011}, which correlates over 14000 fMRI studies with
1300 descriptors. For the 2-head attention model with QDL loss, we
computed the correlation between the demographic variable $d\left(i\right)$
and the context for functional network $f$ from each head $c_{1}\left(f\right)$
and $c_{2}\left(f\right)$ across the test ASD subjects. We analyzed
the US and Yale site as their test accuracy was high ($>75$\%) and
they contained significant heterogeneity for the investigated demographic
variables of age, gender, handedness, and full IQ. We then found the
functional network $f$ that resulted in the largest difference in
correlation values for the 2 heads. The binary mask of the functional
network of interest was then input to Neurosynth to assess neurocognitive
processes associated with different modes of heterogeneity in ASD.
\begin{table}[t]
\begin{centering}
\protect\caption{\label{tab:DS1}DS1 Classification Results (N = 1100, 48.1\% ASD)}
\resizebox{0.999\columnwidth}{!}{%
\begin{tabular}{|c||c|c|c|c||c|c|c|}
\hline 
 & \multicolumn{4}{c||}{Leave-One-Site-Out} & \multicolumn{3}{c|}{Weighted by \# Subjects/Site}\tabularnewline
\cline{2-8} 
Model & ~Mean (Std)~ & ~Mean (Std) ~ & ~Mean (Std)~ & \multicolumn{1}{c||}{~Mean (Std)~} & ~Mean (Std)~ & ~Mean (Std) ~ & ~Mean (Std)~\tabularnewline
 & ACC (\%) & TPR (\%) &  TNR (\%) & AUC & ACC (\%) & TPR (\%) &  TNR (\%)\tabularnewline
\hline 
Orig (LSTM) \cite{Dvornek2017} & 63.4 (0.7) & 60.9 (1.2) & 66.2 (0.5) & 0.695 (0.006) & 65.0 (0.7) & 61.3 (1.3) & 68.4 (1.2)\tabularnewline
\hline 
DFuse \cite{Dvornek2018} & 63.3 (1.2) & 55.7 (3.3) $^{\diamond}$ & 70.7 (2.5) {*} & 0.701 (0.017) & 65.4 (1.3) & 57.7 (3.3) & 72.5 (3.3)\tabularnewline
\hline 
DInit \cite{Dvornek2018a} & 65.4 (0.6) {*} & 60.7 (1.2) & 69.9 (0.6) {*} & 0.709 (0.006) & \textbf{67.1 (0.7)} {*} & 62.6 (2.4) & 71.3 (2.5) {*}\tabularnewline
\hline 
DGA1-C & 64.4 (0.7) & \textbf{62.5 (0.6)} & 66.3 (1.5) & 0.710 (0.009) & 65.9 (0.5) & \textbf{63.5 (1.4)} & 68.1 (0.9)\tabularnewline
\hline 
DGA2-C & 64.3 (1.2) & 56.2 (2.3) $^{\diamond}$ & 71.8 (3.0) {*} & 0.703 (0.006) & 65.8 (1.1) & 57.3 (1.7) $^{\diamond}$ & 73.8 (3.0) {*}\tabularnewline
\hline 
DGA1 & 63.8 (0.9) & 61.5 (2.9) & 66.1 (1.9) & 0.702 (0.009) & 65.7 (1.1) & \textbf{63.5 (3.0)} & 67.7 (2.7)\tabularnewline
\hline 
DGA2 & 64.8 (2.4) & 56.1 (3.8) & \textbf{73.1 (2.1) {*}} & 0.710 (0.011) & 66.3 (1.6) & 57.1 (3.2) $^{\diamond}$ & \textbf{74.8 (2.5)} {*}\tabularnewline
\hline 
DGA2-QDL & \textbf{65.5 (0.8) }{*} & 59.1 (2.3) & 72.0 (2.4) {*} & \textbf{0.711 (0.006)} & 66.8 (0.7) {*} & 60.7 (1.3) & 72.4 (1.9)\tabularnewline
\hline 
\end{tabular}}
\par\end{centering}

{\small{}{*} Significantly different compared to LSTM with no demographic
input ($p<0.05$), with larger mean value.}{\small \par}

{\small{}$^{\diamond}$ Significantly different compared to LSTM with
no demographic input ($p<0.05$), with smaller mean value.}{\small \par}
\end{table}
\begin{table}[t]
\begin{centering}
\protect\caption{\label{tab:DS2}DS2 Classification Results (N = 1035, 48.8\% ASD) }
\resizebox{0.999\columnwidth}{!}{%
\begin{tabular}{|c||c|c|c|c||c|c|c|}
\hline 
 & \multicolumn{4}{c||}{Leave-One-Site-Out} & \multicolumn{3}{c|}{Weighted by \# Subjects/Site}\tabularnewline
\cline{2-8} 
Model & ~Mean (Std)~ & ~Mean (Std) ~ & ~Mean (Std)~ & \multicolumn{1}{c||}{~Mean (Std)~} & ~Mean (Std)~ & ~Mean (Std) ~ & ~Mean (Std)~\tabularnewline
 & ACC (\%) & TPR (\%) &  TNR (\%) & AUC & ACC (\%) & TPR (\%) &  TNR (\%)\tabularnewline
\hline 
Orig$^{\dagger}$ \cite{Heinsfeld2018} & 65 (1.5) & 69 (2.6)  & 62 (2.7) & - & 65.4 (1.3) & 68.1 (2.6) & 62.3 (2.6)\tabularnewline
\hline 
LSTM \cite{Dvornek2017} & 63.6 (0.5) & 55.2 (1.6) & 71.9 (0.6) & 0.709 (0.006) & 65.6 (0.6) & 58.2 (1.7) & 72.7 (0.9)\tabularnewline
\hline 
DFuse \cite{Dvornek2018} & 65.5 (0.9) {*} & 57.1 (0.6) & 73.5 (1.6) & 0.713 (0.006) & 67.2 (0.6) & 61.2 (1.2) & 72.8 (1.0)\tabularnewline
\hline 
DInit \cite{Dvornek2018a} & 65.8 (0.8) {*} & 58.1 (0.4) & 72.9 (1.4) & 0.720 (0.009) & \textbf{67.5 (1.1)} {*} & 61.8 (1.6) {*} & 72.9 (3.2)\tabularnewline
\hline 
DGA1-C & 65.6 (1.7) {*} & 61.1 (1.6) & 69.6 (1.1) & 0.713 (0.011) & 66.8 (1.6) & \textbf{64.1 (2.0)} {*} & 69.3 (1.9)\tabularnewline
\hline 
DGA2-C & 65.8 (0.9) {*} & 52.6 (2.4) & \textbf{78.3 (1.7) }{*} & 0.719 (0.009) & 67.2 (1.2) {*} & 55.9 (2.4) & \textbf{78.0 (0.8) }{*}\tabularnewline
\hline 
DGA1 & 66.1 (1.5) {*} & \textbf{61.3 (2.5)} {*} & 70.4 (1.4)  & 0.719 (0.011) & 67.4 (1.7) {*} & 63.6 (2.3) {*} & 70.9 (1.7) \tabularnewline
\hline 
DGA2 & 65.5 (1.0) {*} & 54.3 (1.5)  & 76.5 (1.4) {*} & 0.716 (0.015) & 67.1 (1.4) & 57.6 (1.3) & 76.1 (2.3) {*}\tabularnewline
\hline 
DGA2-QDL & \textbf{66.4 (0.4)} {*} & 58.0 (1.9) {*} & 74.2 (2.0) & \textbf{0.722} (0.006) & 67.4 (0.5) {*} & 61.3 (1.7) {*} & 73.1 (1.9)\tabularnewline
\hline 
\end{tabular}}
\par\end{centering}

{\small{}$^{\dagger}$ Values taken from the literature, reflecting
one round of LOSO CV.}{\small \par}

{\small{}{*} Significantly different compared to LSTM with no demographic
input ($p<0.05$).}{\small \par}
\end{table}
\begin{table}[t]
\begin{centering}
\protect\caption{\label{tab:DS3}DS3 Classification Results (N = 860, 46.1\% ASD)}
\resizebox{0.999\columnwidth}{!}{%
\begin{tabular}{|c||c|c|c|c||c|c|c|}
\hline 
 & \multicolumn{4}{c||}{Leave-One-Site-Out} & \multicolumn{3}{c|}{Weighted by \# Subjects/Site}\tabularnewline
\cline{2-8} 
Model & ~Mean (Std)~ & ~Mean (Std) ~ & ~Mean (Std)~ & \multicolumn{1}{c||}{~Mean (Std)~} & ~Mean (Std)~ & ~Mean (Std) ~ & ~Mean (Std)~\tabularnewline
 & ACC (\%) & TPR (\%) &  TNR (\%) & AUC & ACC (\%) & TPR (\%) &  TNR (\%)\tabularnewline
\hline 
Orig$^{\dagger}$ \cite{Abraham2017} & 63.6 (6.2) & 59.8 (10.3) & 66.7 (12.8) & - & - & - & -\tabularnewline
\hline 
LSTM \cite{Dvornek2017} & 63.8 (0.4) & 50.3 (1.9) & 75.5 (1.4) & 0.694 (0.012) & 65.3 (0.6) & 53.9 (2.3) & 75.0 (1.9)\tabularnewline
\hline 
DFuse \cite{Dvornek2018} & 65.6 (1.6) & 52.7 (1.8) & 76.4 (3.0) & 0.714 (0.007) & 67.1 (0.8) {*} & 56.9 (2.5) & 75.8 (1.4)\tabularnewline
\hline 
DInit \cite{Dvornek2018a} & 64.5 (1.2) & 50.5 (2.2) & 76.4 (1.9) & 0.702 (0.013) & 66.3 (0.8) & 55.5 (2.4) & 75.5 (2.4)\tabularnewline
\hline 
DGA1-C & 65.5 (1.2) {*} & \textbf{55.3 (1.6)} {*} & 74.5 (2.1) & 0.708 (0.010) & 66.8 (0.9) {*} & 59.0 (2.4) {*} & 73.5 (3.4)\tabularnewline
\hline 
DGA2-C & 65.9 (1.6) & 52.6 (1.7) & 78.6 (2.0) & \textbf{0.717 (0.014)} & 67.2 (1.2)  & 55.2 (1.8) & 78.6 (1.5)\tabularnewline
\hline 
DGA1 & 65.8 (0.1) {*} & \textbf{55.3 (1.1)} {*} & 75.2 (1.2) & 0.712 (0.006) & 66.8 (0.7) {*} & \textbf{59.1 (1.8)} {*} & 73.3 (1.5)\tabularnewline
\hline 
DGA2 & \textbf{66.8 (1.0)} {*} & 51.2 (2.1) & \textbf{80.0 (2.7) }{*} & 0.714 (0.005) & \textbf{68.0 (1.0)} {*} & 54.0 (2.5) & \textbf{80.0 (2.7)} {*}\tabularnewline
\hline 
DGA2-QDL & 66.0 (1.1) & 53.5 (1.4) & 76.8 (3) & 0.709 (0.006) & 67.0 (0.9) {*} & 57.6 (2.9) & 75.2 (3.3)\tabularnewline
\hline 
\end{tabular}}
\par\end{centering}

{\small{}$^{\dagger}$ Values obtained from corresponding author of
}\cite{Abraham2017}{\small{}, reflecting one round of LOSO CV.}{\small \par}

{\small{}{*} Significantly different compared to LSTM with no demographic
input ($p<0.05$).}{\small \par}
\end{table}

\subsection{Classification Results}

Classification results for each dataset are summarized in Tables \ref{tab:DS1}-\ref{tab:DS3}.
The results using the method from the original study for DS1 and published
in the original study for DS2 and DS3 use only fMRI data and are shown
in the first entry. We notice that generally, the fusion model DFuse
and LSTM initialization model DInit do not perform significantly differently
from the baseline LSTM model, particularly for DS3. The DGA-based
models that use the context alone as the input to the FC (DGA1-C and
DGA2-C) tend to perform about the same (DS1) or better (DS2 and DS3)
than the non-DGA models. Adding in the residual connection for DGA1
and DGA2 results in similar (DS1 and DS2) or better (DS3) results
than the DGA-C models. Finally, the DGA2-QDL model resulted in the
top performance for DS1 and DS2 as measured by accuracy and AUC. 

To better understand the performance over all the datasets, we scored
each model by the number of performance measures that significantly
improved over the baseline LSTM, minus the number of measures that
significantly worsened compared to baseline, plus the number of top
ranked measures. The models ranked in order of increasing performance
was then DFuse, DGA2-C, DInit, DGA1-C, DGA1, DGA2-QDL, DGA2. Thus,
DGA-based models generally performed better than other demographic
models; 2-headed attention was generally better than 1; and the proposed
residual connection for using the context as a bias to the LSTM outputs
generally performed better than using the context alone. The reason
for DGA2-QDL's lower ranking is due to the performance on DS3; we
posit that the lower number of subjects in this dataset led to less
heterogeneity, thus making it difficult to find two disparate attention
modes, which QDL is trying to recover by minimizing the projection
space similarity. 
\begin{figure}[H]
\begin{centering}
\hspace*{-6bp}\subfloat[US site, Age]{\protect\includegraphics[height=0.11\textheight]{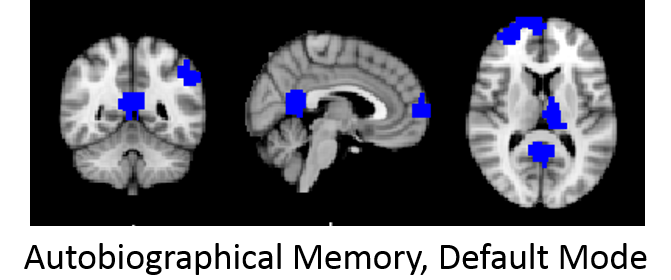}

\vspace{-20bp}
}\hspace*{0bp}\subfloat[US site, IQ]{\protect\includegraphics[height=0.11\textheight]{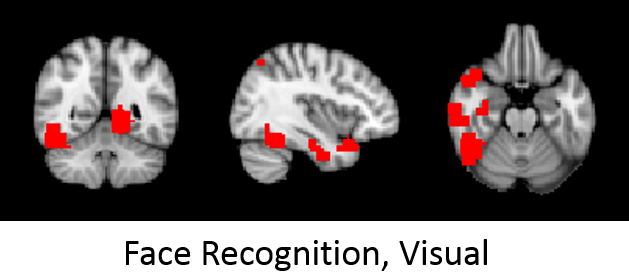}\vspace{-10bp}
}
\par\end{centering}

\begin{centering}
\subfloat[Yale site, Gender]{\protect\includegraphics[height=0.11\textheight]{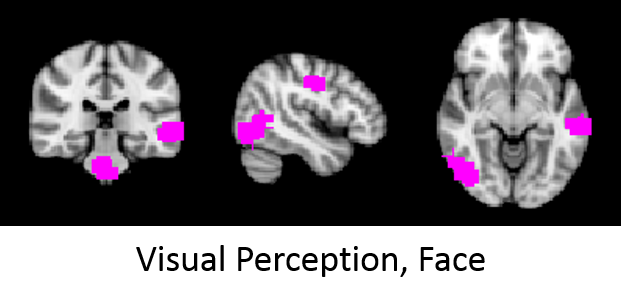}\vspace{-10bp}
}~~\subfloat[Yale site, Handedness]{\protect\includegraphics[height=0.11\textheight]{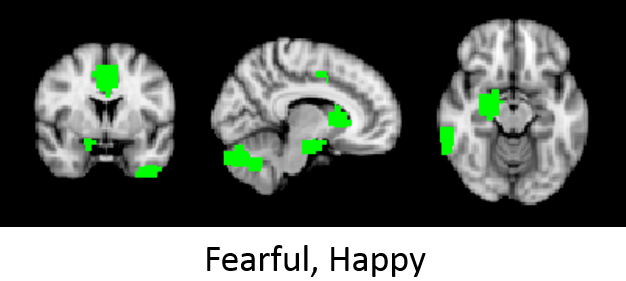}\vspace{-10bp}
}
\par\end{centering}

\protect\caption{\label{fig:neurosynth}Functional networks from the DGA2-QDL model
trained on DS2 which had largest difference between the correlations
of the listed demographic variable with the two attention measures
$c_{1}\left(f\right)$ and $c_{2}\left(f\right)$ for ASD subjects
in the listed test site. The top associated cognitive functions decoded
by Neurosynth for each network are shown. }
\end{figure}

\subsection{Demographic-guided Heterogeneity of Functional Processing}

We explored the functional networks from the best model for DS2, DGA2-QDL,
that corresponded to the most diverse outputs by the 2 attention heads.
These different modes of the model's response to a functional network
may correspond to potentially different mechanisms of ASD pathophysiology.
Resulting functional networks and the top 2 associated Neurosynth
cognitive terms are shown in Fig. \ref{fig:neurosynth}. The functional
networks highlight regions that are often associated with ASD (e.g.,
Fig.~\ref{fig:neurosynth}(b) and (c), visual perception and face
processing \cite{Kaiser2010}), and are also potentially associated
with the demographic variable of interest (e.g., Fig.~\ref{fig:neurosynth}(a),
default mode network changes with age \cite{Fair2008}).

\section{Conclusions}

We have presented a novel demographic-guided attention mechanism for
modeling the heterogeneity in neuropathophysiology of ASD. We achieved
higher ASD classification performance on several ABIDE datasets and
preprocessing conditions under a leave-one-site-out cross-validation
framework, demonstrating improved generalization to data from new
imaging sites. The success of having multiple attention modes for
modeling the different neural mechanisms associated with ASD may help
partially explain some of the conflicting results in the ASD literature
(e.g., hyper- vs. hypo-connectivity), as our classification models
improve once we account for the heterogeneity of the disorder.

\clearpage{}

\bibliographystyle{splncs04}
\bibliography{mlmi2020_ref}

\end{document}